\documentclass[letterpaper, 10 pt, conference]{ieeeconf}  

\IEEEoverridecommandlockouts                              

\usepackage{gradient-text}

\title{\LARGE \bf

\textsf{\gradientRGB{OpenNav}{254,50,254}{15,224,238}}: Open-World Navigation \\ with Multimodal Large Language Models
\vspace{-0.5em}}

\author{Mingfeng Yuan$^{1}$, Letian Wang$^{1}$ and Steven L. Waslander$^{1}$ 
\thanks{Authors are with the University of Toronto Institute for Aerospace Studies and the University of Toronto Robotics Institute, Toronto, Canada
	{\tt\small\{mingfeng.yuan,letian.wang, steven.waslander\}@robotics.utias.utoronto.ca}}
}

\def\BibTeX{{\rm B\kern-.05em{\sc i\kern-.025em b}\kern-.08em
    T\kern-.1667em\lower.7ex\hbox{E}\kern-.125emX}}
\usepackage{balance}
\usepackage{amsmath,amsfonts}
\usepackage{array}
\usepackage{textcomp}
\usepackage{stfloats}
\usepackage{url}
\usepackage{verbatim}
\usepackage{graphicx}
\usepackage{cite}
\usepackage[dvipsnames]{xcolor}
\usepackage{caption, subcaption}
\usepackage{float}
\usepackage{diagbox}
\usepackage{caption}
\usepackage{booktabs}
\usepackage{epstopdf}
\usepackage{bm}
\usepackage{hyperref}
\usepackage{multirow}
\usepackage{cuted}
\usepackage{booktabs} 
\usepackage{afterpage}  
\usepackage{xcolor}

\usepackage{algorithm}
\usepackage{algpseudocode}
\usepackage{amsmath}

\begin{document}

\makeatletter
\let\@oldmaketitle\@maketitle
\renewcommand{\@maketitle}{\@oldmaketitle
\centering
\vspace{0.5em}
\includegraphics[width=0.85\linewidth, trim={1.1cm 0 0.8cm 0}]{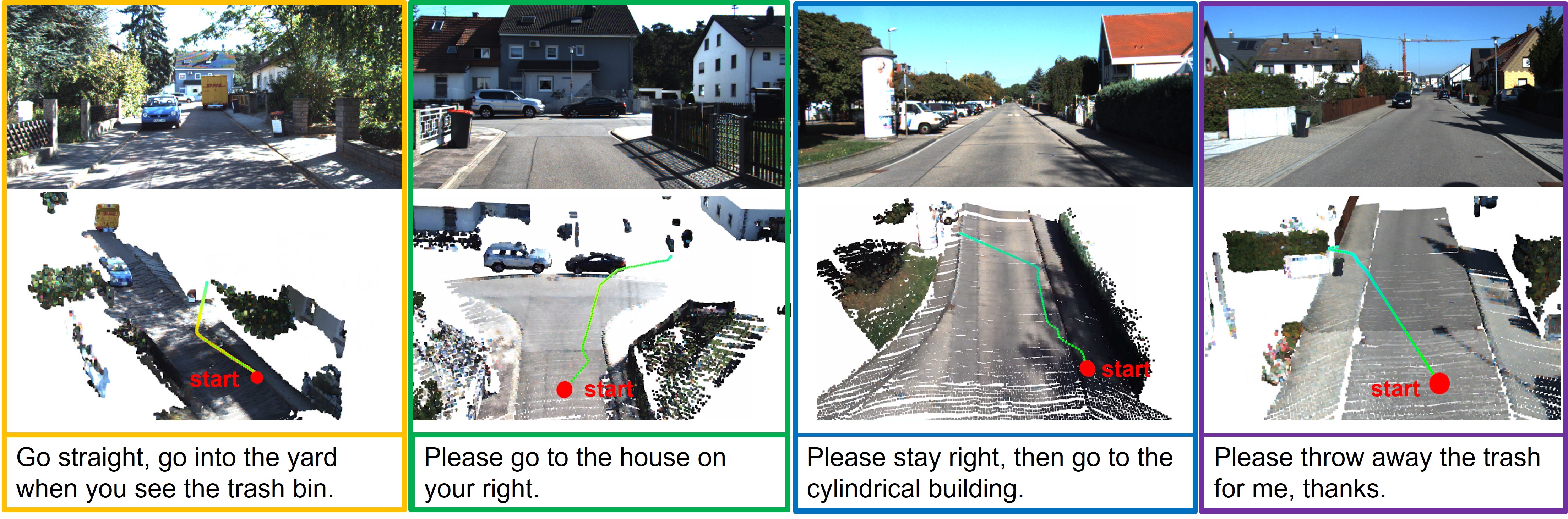}
\captionof{figure}{
Given a free-form language instruction and sensor observations, OpenNav is capable of generating a dense sequence of instruction-following and scene-compliant robot waypoints in a zero-shot manner for open-world navigation, effectively handling open-set objects and open-set instructions without relying on in-context examples or pre-trained skills.
}
\label{fig:sample}
\vspace{-0.98em}
}
\makeatother

\maketitle\
\thispagestyle{empty}
\pagestyle{empty}

\begin{abstract}
Pre-trained large language models (LLMs) have demonstrated strong common-sense reasoning abilities, making them promising for robotic navigation and planning tasks. However, despite recent progress, bridging the gap between language descriptions and actual robot actions in the open-world, beyond merely invoking limited predefined motion primitives, remains an open challenge. In this work, we aim to enable robots to interpret and decompose complex language instructions, ultimately synthesizing a sequence of trajectory points to complete diverse navigation tasks given open-set instructions and open-set objects. We observe that multi-modal large language models (MLLMs) exhibit strong cross-modal understanding when processing free-form language instructions, demonstrating robust scene comprehension. More importantly, leveraging their code-generation capability, MLLMs can interact with vision-language perception models to generate compositional 2D bird-eye-view value maps, effectively integrating semantic knowledge from MLLMs with spatial information from maps to reinforce the robot’s spatial understanding. To further validate our approach, we effectively leverage large-scale autonomous vehicle datasets (AVDs) to validate our proposed zero-shot vision-language navigation framework in outdoor navigation tasks, demonstrating its capability to execute a diverse range of free-form natural language navigation instructions while maintaining robustness against object detection errors and linguistic ambiguities. Furthermore, we validate our system on a Husky robot in both indoor and outdoor scenes, demonstrating its real-world robustness and applicability. Supplementary videos are available at \textcolor{magenta}{https://trailab.github.io/OpenNav-website/}


\end{abstract}
\section{INTRODUCTION}
With the rapid advancement of Vision-Language Models (VLMs) in recent years, their ability to align natural language with visual information and perform semantic reasoning has significantly reshaped robot perception and human-robot interaction. Consequently, embodied intelligence has emerged as a prominent research area\cite{ramrakhya2022habitat}. 
However, most of these advancements have limitedly remained in structured scenarios, such as indoor fixed tabletop manipulation tasks or indoor object goal navigation tasks, involving a few predefined closed-set objects and instructions. When we deploy and unleash robots to the open-world settings, such as outdoor navigation, 
they must contend with environments populated by \textbf{open-set objects with unstructured layout}, resulting in significantly greater complexity and environmental uncertainty. 
Moreover, \textbf{open-set instructions of free-form} in real applications introduce additional challenges by requiring flexible and hierarchical environmental understanding, as different instructions may demand varying levels of spatial, semantic, and relational reasoning - from distinguishing similar objects to identifying specific landmarks to understanding abstract spatial relationships across large, unstructured outdoor spaces.
In such settings, existing methods\cite{parakh2024lifelong} face significant challenges as they rely entirely on object detectors for environmental descriptions and use LLMs for reasoning in the linguistic domain, leading to the loss of crucial spatial and subtle information essential for navigation tasks. As a result, they struggle to handle the richness and variability of open-world applications.

In addition to the open-world challenges, a fundamental challenge persists in translating MLLM's powerful language and common-sense reasoning capabilities into physical execution for robotic control. Given the inherent lack of exposure of LLMs and VLMs to physical interaction data and geometry information during training, existing efforts have largely assumed that these models are not well suited for hardware-dependent low-level control\cite{kwon2024language}, and thus, many works adopt LLMs as high-level planners, neglecting low-level robot motion control or relying on predefined motion primitives (e.g. point-to-point navigation), which compromises the flexibility of language instructions in controlling the robot's motion behavior. In contrast, a significant contribution on planning with dynamic behaviors is demonstrated in the LQR-RRT$^*$ framework, highlighting the importance of integrating control feasibility into trajectory synthesis\cite{zhong2023motion}. The emergence of VoxPoser\cite{huang2023voxposer} has also demonstrated that LLMs excel at inferring affordances and constraints when provided with grasping task instructions. However, many zero-shot frameworks, including VoxPoser, relied heavily on providing in-context examples to the LLM input, making them unsuitable for open-world navigation tasks. Given these challenges, our contributions are threefold: 
\begin{itemize}
 \item We introduce \textbf{OpenNav}, a zero-shot vision-language navigation framework that, to the best of our knowledge, is the first to directly generate trajectories using a MLLM for outdoor navigation given open-set instructions and open-set objects, without relying on pre-trained skills, motion primitives, or in-context examples. 
\item We propose a multi-expert system that integrates state-of-the-art MLLMs with an open-vocabulary perception system to enhance robotic scene comprehension. Through the use of a single task-agnostic prompt and a multimodal interface between the MLLM and the perception system, our framework significantly improves robustness against the uncertainties under open-set objects and language instructions. By combining the reasoning, code generation, function-calling capabilities of MLLMs with classical planning techniques, our approach harnesses the benefits of both human-like reasoning and geometry-compliant trajectory synthesis.
\item Our OpenNav is evaluated on both AVDs and a ground robot, highlighting its effectiveness in vision-language navigation and embodied intelligence research.
\end{itemize}

\section{Related works}
\subsection{Open-Set Perception and Planning}
Early works on object detection and semantic 3D mapping primarily relied on deep learning-based approaches and labeled datasets\cite{shi2024city} to obtain spatial representations of objects in the environment (such as occupancies and point clouds). However, their performance was largely constrained by the richness of the training data, as these models were trained in a closed-set of object classes and specific scene datasets under a supervised learning paradigm. This limitation makes it challenging for such models to recognize unseen object classes in complex and open environments. Recent VLMs trained on internet-scale data enable open-vocabulary 2D and 3D understanding. In robotics applications, recent approaches commonly utilize models such as SAM and CLIP to generate segmented object point clouds with associated detection labels and feature embeddings\cite{gu2024conceptgraphs}. Additionally, some works structure spatially-grounded textual scene descriptions from point cloud data\cite{parakh2024lifelong}. While effective in simple and small-scale environments (e.g., tabletop scenes), this approach becomes insufficient for complex and large-scale open environments where fine-grained spatial relationships are crucial. This approach is analogous to a person attempting to plan actions in an environment solely based on verbal descriptions provided by another person while keeping their eyes closed. Unlike the existing works that decouple open-vocabulary object detection from LLM-based planning, we leverage recent advances in MLLMs and adopt a multi-expert system framework where a single model is responsible for both scene comprehension and decision-making. In our approach, an open-vocabulary perception system (OVPS) is employed to obtain detailed captions of segmented objects as well as 3D semantic-geometric map, thereby enhancing the MLLM’s ability to interpret ambiguous language instructions and perform reasoned decision-making in navigation tasks.

\subsection{Vision-Language Navigation}
Vision-language navigation (VLN) is a core task in embodied intelligence, requiring agents to navigate complex environments using visual cues and natural language instructions. The field has made significant progress in recent years. Early research on discrete navigation in simulators like Matterport3D focused on teleportation-based movement between predefined nodes\cite{zhou2024navgpt}. With the rise of pre-trained LLMs, many studies leverage large-scale models\cite{majumdar2020improving} and pre-training techniques\cite{guhur2021airbert} to enhance navigation, but they primarily address high-level decision-making, neglecting low-level motion control.
Recent work\cite{zhang2024navid} has shifted to continuous environments, where agents navigate using mid-level actions (e.g., move forward, rotate in place), as in VLN-CE with Habitat\cite{ramrakhya2022habitat}. To bridge the gap between discrete and continuous navigation, some approaches employ waypoint models to predict candidate positions\cite{hong2022bridging}, improving performance but suffering from limited generalization and lack of motion planning or obstacle avoidance.  This study explores a new paradigm, leveraging MLLMs for trajectory generation, unifying high-level decision-making and motion planning at the trajectory level while incorporating obstacle avoidance for more effective navigation.


\subsection{Code as Policy}
Prior work has leveraged LLMs as planners for robotic manipulation, notably SayCan\cite{ahn2022can}, which relies heavily on in-context examples at inference time\cite{huang2023voxposer}. While our approach shares similarities with Kwon et al.\cite{kwon2024language}, their focus remains on manipulation tasks. In contrast, we employ a single, task-agnostic prompt—free from in-context examples or predefined motion primitives—to enable generalizable navigation. Our model autonomously invokes APIs to retrieve task-relevant information, generates executable trajectory code, and can detect, debug, and re-execute upon failure, enhancing adaptability and robustness.

\section{Method}
\begin{figure*}[t] 
	\centering
	\includegraphics[width=0.9\linewidth]{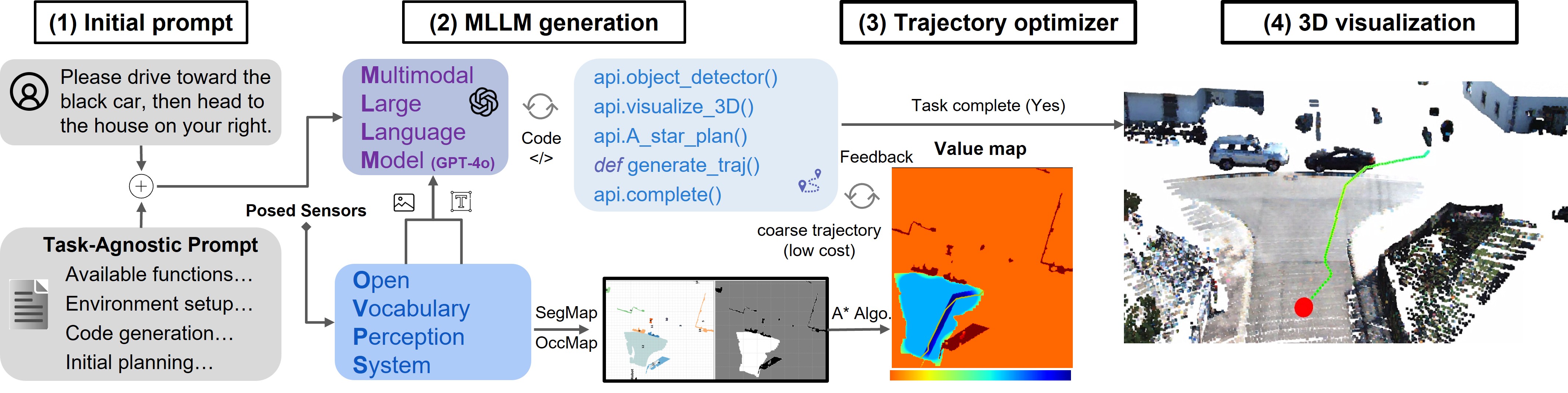}
	\caption{Overview of OpenNav. Given the posed RGB-Lidar observation of the environment and an open-set free-form language instruction, 1) we leverages task-agnostic prompts to enable zero-shot generalization and adaptability to varied instructions; 2) MLLM generates code, which interacts with OVPS, to produce open-set multimodal scene perception outputs, and 2D bird-eye-view (BEV) value map (consists of a semantic map and occupancy map) grounded in the operation environment. 3) MLLM synthesizes a human-like coarse trajectory based on instructions, scene understanding, and its reasoning capabilities. The generated BEV value map then serves as the objective function for the motion planner, which refines the trajectory to ensure geometry-compliant navigation. Please see Fig~\ref{fig:perception} for detailed pipeline, inputs, and outputs of the OVPS.}
    \vspace{-0.4cm}
	\label{fig:framework}
\end{figure*}

In this study, we investigate the extent to which a pretrained MLLM can address navigation tasks with open-set instruction and open-set object in a zero-shot manner, without relying on in-context examples or predefined motion primitives. To explore this problem, we present OpenNav, a zero-shot VLN framework that integrates an open vocabulary perception system, a high-level planning empowered by MLLM, and a mid-level trajectory optimizer. As illustrated in Fig. \ref{fig:framework}, our approach leverages the inherent multimodal understanding, code generation, reasoning, and planning capabilities of MLLMs to predict dense trajectories that enable successful task execution. The pipeline of the algorithm can be found in \textbf{Algorithm} \ref{pipeline}. The following sections provide a detailed description of our framework, where we will define the key challenges and introduce our proposed solution. We first introduce how the system works as a whole in Section~\ref{sec: method - system}, and then elaborate the developed open vocabulary perception system in Section~\ref{sec: perception}.

\subsection{Open-Set Zero-Shot Vision-Language Navigation}
\label{sec: method - system}
\subsubsection{\textbf{Open-Set Instructions}}
A navigation task is specified through a free-form language command $l$ (e.g., \textit{"Go to the house on your right"}). To successfully execute such tasks, the robot must possess the following capabilities: (a) comprehension and reasoning about the instruction; (b) decomposition of the high-level task into subtasks, such as object detection, trajectory shape planning, and trajectory generation via code synthesis; (c) sequential execution of each subtask; and (d) detection of task failure and re-planning accordingly. The primary challenge of handling open-set instructions lies in the inability to predefine task execution using in-context examples or rule-based methods, thereby requiring strong zero-shot capabilities. To address this, we design a single \textbf{task-agnostic prompt}, $p$, that enables the robot to generalize across diverse outdoor navigation tasks, ranging from specific (e.g., \textit{"Go to the red car."}), to abstract (e.g., \textit{"Throw away the trash."}), to directional (e.g., \textit{"Drive toward the black car, then head to the house."}). This system prompt comprises five key components: 
\begin{enumerate}
    \item \textbf{Available Functions}, listing callable APIs such as \texttt{det\_object()}, \texttt{A\_star\_plan(start, end)} for open-set perception and value-map-based trajectory planning, and \texttt{visual\_3D(traj)} for trajectory visualization.
    \item \textbf{Environment Description}, detailing the coordinate system, robot pose, and sensor configurations.
    \item \textbf{Collision Avoidance Guidance}, instructing the MLLM to identify potential obstacles and traversable areas before execution.
    \item \textbf{Initial Planning}, prompting the model to perform step-by-step reasoning and cross-modal understanding of sensory observations. This stage requires the MLLM to complete the previously mentioned (a) and (b) tasks based on system prompt and the given tasks.
    \item \textbf{Code Generation}, requiring MLLM to either call existing APIs or generate code to produce the trajectory in segments, which are then subsequently concatenated into a complete trajectory. This part is designed to complete the previously mentioned (c) and (d) tasks.
\end{enumerate}
This task-agnostic prompt along with the navigation task in free-form language, is provided as input to the MLLM as shown in Fig. \ref{fig:framework}, where we utilize ChatGPT-4o, one of the most advanced proprietary models. 

\subsubsection{\textbf{Open-Set Objects}}
In real-world navigation tasks, a diverse range of objects may be encountered, and target objects can be described in various ways: (1) by their inherent category, such as \textit{house}; (2) through feature-based descriptions, such as \textit{cylindrical building}, which reference shape, appearance, or other attributes; or (3) via functional descriptions, such as \textit{throw away the trash}, where the most relevant object would be a trash bin. To handle such variability, we employ a multi-expert object perception system. The first component is an OVPS, composed of a set of open-source vision-language perception models, responsible for object detection, segmentation, caption generation, and extracting object attributes such as position and size, the implementation details of which will be introduced in the the coming Section~\ref{sec: perception}. The second expert model in the system leverages the multimodal understanding and reasoning capabilities of a MLLM. The output of the OVPS is an annotated RGB image, along with textual descriptions of each object's attributes including captions, center pose, dimensions, and the nearest reachable point (NRP) on each drivable region to the target object (formatted as a dictionary $\{Reg.: Point, ...\}$). This information, combined with the prompts from the previous section, serves as inputs to the MLLM and constitutes the MLLM’s environmental perception.

\subsubsection{\textbf{Instruction Grounding: Zero-Shot Trajectory Synthesis via MLLM}}
\label{sec: VLT}
Interpreting and executing free-form language instructions is a fundamental yet challenging problem in embodied intelligence. Given the environmental perception described in previous subsections, the MLLM begins by understanding and reasoning about the task. For instance, given the instruction \textit{``drive toward the black car, then head to the house on your right"} as shown in Fig. \ref{fig:sample}, the MLLM leverages its semantic understanding and task planning capabilities to identify relevant targets such as the \textit{house}, drivable surfaces, and obstacles along the route. Specifically, when multiple similar objects exist in the environment, the model must further determine the relevant and target objects using spatial relations mentioned in the instruction (e.g., \textit{"on your right"}).
After identifying the target and relevant objects, it then infers the trajectory shape, by first selecting an intermediate waypoint near relevant objects, and then choosing the nearest detected drivable surface point to the target object. For example, for the above instruction, the model will first driving toward the black car and then making a right turn toward the final target house.


Specifically, to generate trajectories aligned with linguistic instructions, we exploit the powerful code generation abilities of MLLMs to bypass their inadequacy in directly generating continuous trajectories. 
Given environmental observations, the MLLM first infers the trajectory shape by reasoning about the task and then synthesizes a Python script to generate the trajectory, specifying the start point, intermediate waypoints, and the endpoint. This script is executed in a Python interpreter, and if any errors occur, the MLLM automatically debugs the script based on terminal error messages and re-executes it until a valid trajectory is produced. This process ultimately yields a dense 3D trajectory sequence. 
\begin{algorithm}[!t]
\begin{algorithmic}[1]
\Require task-agnostic prompt $p$, task instruction $l$, and observations \{text: (\textit{cap}, \textit{pos}, \textit{dim}, \textit{nrp}); visal: \textit{I}\}

\State $prompt \gets p \oplus l \oplus I$
\Comment{concatenate}
\State $task\_completed \gets \text{False}$
\While{$task\_completed = \text{False}$}
    \State $output \gets \text{MLLM}(prompt)$
    \State $prompt \gets \text{None}$
    \If{$output$ contains code}
        \State \textbf{try} $exec(output)$ \Comment{extract code}
        \State \textbf{except} Exception then
        
        \State \indent$prompt \gets$ error message
    \Else
        \If{$detect\_object()$ is called}
            \State $prompt \gets (cap, pos, dim, nrp) \oplus \textit{I}$
        \ElsIf {$(\textit{reg}, \textit{traj})$ are avaliable}
            \State $val\_map \gets occ\_map \oplus val\_map(traj, reg)$ 
            \State $final\_t \gets a\_star\_plan(val\_map)$
            \State $visual\_3D(final\_t)$
            \State \Comment {\textit{final\_t}: \textbf{final t}rajectory;}
            \State \Comment{\textit{nrp}: \textbf{n}earest \textbf{r}eachable \textbf{p}oints}
            \State \Comment{\textit{traj}: \textbf{tra}jectory generated by code;} 
            \State \Comment{\textit{reg}: identified  drivable \textbf{reg}ion index;}
        \EndIf
    \EndIf
\EndWhile
\end{algorithmic}
\caption{Open-World Navigation Pipeline}\label{pipeline}
\end{algorithm}

\subsubsection{\textbf{Geometry-Compliant Trajectory Refinement}}
\label{sec: OpenNav}
Despite MLLM's reasoning capabilities, we observe that even state-of-the-art proprietary models struggle to generate precise, smooth, and collision-free trajectories. To address this issue, we incorporate traditional path-planning algorithms, such as \textbf{A*}, which efficiently generate collision-free trajectories based on geometric maps but inherently prioritize shortest trajectory and lack semantic understanding of the environment. To align trajectory generation with linguistic instructions, we propose a framework that integrates the MLLM with \textbf{A*} planning, serving as trajectory optimizer. 
Specifically, the trajectory generated by MLLM is regarded as a coarse trajectory segment aligned with the task description, which is then projected onto a 2D bird-eye-view (BEV) value map, where the traversed regions (considering a predefined radius around each point) are assigned a lower cost than the surrounding drivable surface areas, ensuring alignment between language-based intent and motion planning. Note that the initial value map is jointly initialized using the occupancy map and the semantic map. Regions with a value of 1 in the occupancy map are assigned a high cost in the value map to ensure collision avoidance. Additionally, to control navigable areas (e.g., distinguishing between paved roads and sidewalks), the value map assigns a slightly higher cost to navigable regions than the areas covered by the coarse trajectory, while other ground regions are assigned the same high cost as occupied areas. Please refer to the Section~\ref{sec: perception} for details.
\textbf{A*} then generates a refined, collision-free trajectory based on the updated BEV value map, ensuring alignment with linguistic guidance. This approach also mitigates the limitations of traditional planners in understanding environmental semantics. For instance, avoiding a flooded area is challenging for \textbf{A*} alone, as geometric maps lack this information, whereas an MLLM can infer occupancy conditions from a semantic map to guide trajectory generation. Finally, we apply B-spline smoothing to the trajectory and visualize it in a reconstructed 3D map for validation and success rate evaluation.

\begin{figure*}[t] 
	\centering
	\includegraphics[width=0.9\linewidth]{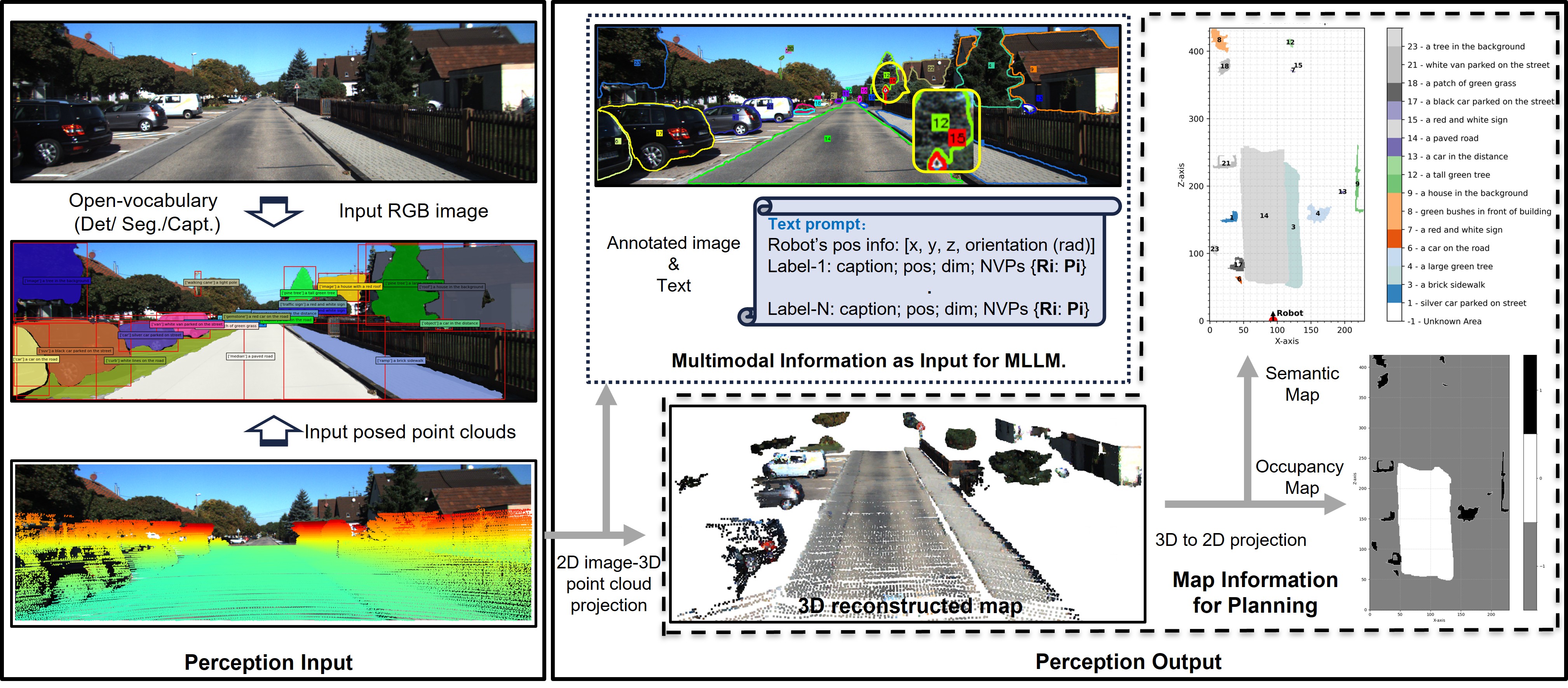}
	\caption{Our open vocabulary perception system sequentially performs detection, segmentation, and object caption generation. Combined with 3D point clouds, the system will generate 1) multimodal observations for VLN, including text prompts and image prompts for MLLMs, 2) 3D reconstructed map, as well as 2D occupancy and semantic maps for trajectory refinement.}
	\label{fig:perception}
    \vspace{-1.5em}
\end{figure*}
\subsection{Open Vocabulary Perception System for VLN}
\label{sec: perception}
Our open vocabulary perception system (OVPS) consists of two components: an open vocabulary object perception module and a map reconstruction module, built upon the OpenGraph framework\cite{deng2024opengraph}. This system takes as input RGB images, their corresponding 3D LiDAR point clouds with pose, and outputs two types of information: 1) the open vocabulary object perception module generates multimodal inputs (textual and visual environment information) for MLLMs to conduct scene understanding and task reasoning; 2) the map reconstruction module generates the semantic and occupancy map for trajectory planning and 3D visualization. The overall framework of our perception system is shown in Fig. \ref{fig:perception}, and in the following, we will introduce details of these two modules and outputs.

\textbf{Open Vocabulary Object Perception.} Unlike previous approaches, which primarily rely on text prompts extracted from OVPS to describe scene information as inputs for LLMs in downstream planning tasks, we argue that this approach loses significant visual details that could critically impact a robot’s task success rate, particularly in task-agnostic and open-world environments. Specifically, human-specified tasks often involve nuanced object descriptions, such as color, shape, size, functionality, and spatial relationships, which are challenging to fully represent using formatted textual descriptions alone. In contrast, visual information inherently captures these details. Thanks to the emergence of MLLMs, we can now bridge the gap between the perception and planning modules more effectively, 
where we leverage the perception system to provide both textual and visual information. Specifically, we generate text-based prompts containing detected object captions, global position, and sizes, while simultaneously assigning a unique numerical identifier to each detected object. Additionally, we provide annotated RGB images, in which object contours are outlined with distinct colors and labeled with the corresponding numerical identifiers found in the textual descriptions, aiding MLLMs in differentiating detected objects.

Specifically, at time \( t \), given an input RGB image \( I^{(t)} \), we first apply the Recognize Anything Model (RAM)\cite{zhang2024recognize} , denoted as \( \operatorname{RAM}(\cdot) \), to identify the categories present in the image. With its open-set recognition capability, RAM can detect a wide range of common object categories. Next, the recognized open-vocabulary category names, along with the original image, are passed into the Grounding DINO model\cite{liu2024grounding}, represented as \( \operatorname{GD}(\cdot, \cdot) \), for open-set object detection. This step generates object detection bounding boxes, which serve as essential inputs for the TAP (Tokenize Anything via Prompting) model\cite{pan2024tokenize}, denoted as TAP \( (\cdot, \cdot) \). Guided by the detected bounding boxes, the TAP model performs segmentation and captioning of the primary objects within the image, producing a set of segmentation masks \( \{m_i^{(t)}\}_{i=1,,, m} \) and corresponding textual descriptions \( \{c_i^{(t)}\}_{i=1,,, m} \) for the frame \( I^{(t)} \). The entire process can be formalized as 
\begin{equation}
\left\{m_i^{(t)}, c_i^{(t)}\right\}=\operatorname{TAP}\left(I^{(t)}, \operatorname{GD}\left(I^{(t)}, \operatorname{RAM}\left(I^{(t)}\right)\right)\right)
\end{equation}
where the specific models used in this pipeline can be replaced with alternative models that provide similar functionalities.

\textbf{BEV Value Map Construction.} Recent studies indicate that current pre-trained MLLMs struggle with spatial reasoning due to a lack of depth and geometry information during training. However, we argue that modern robotic platforms are commonly equipped with RGB-D cameras or 3D LiDAR, making it straightforward to obtain object geometry information. Therefore, to generate trajectories that comply with the scene geometry, our perception system generates a 2D BEV value map, including a 2D occupancy map for collision avoidance and a 2D semantic map for class information. 
To initialize the value map, we employ a multisensor calibration and fusion method to project the 3D LiDAR point cloud \(C^{(t)}\) onto a 2D image plane. The projection process extracts object-specific 3D point clouds \(\mathbf{p}_i^{(t)}\) aligned with their corresponding masks \(m_i^{(t)}\):

\begin{equation}
\mathbf{p}_i^{(t)} = \left\{ l_k \mid M_K M_T l_k \in m_i^{(t)}, l_k \in C^{(t)} \right\},
\end{equation}
where \( l_k \) is a LiDAR point, \( M_K \) is the intrinsic camera matrix, and \( M_T \) is the extrinsic transformation matrix for LiDAR-camera alignment. 
The cost values within the value map are subsequently updated using the trajectory generated by MLLM to align with linguistic instructions, as previously discussed in Section~\ref{sec: method - system}.


\section{Experimental Setup}
Due to the lack of outdoor benchmarks for evaluating the performance of ground robots in VLN tasks, we leverage existing AVDs such as SemanticKITTI. This dataset provides rich real-world outdoor scenarios, along with 2D RGB images, 3D LiDAR point clouds, and corresponding pose information, making it an ideal choice for our experiments. Our proposed framework directly generates trajectories in the format of [x, y, z, heading]. To validate the effectiveness of our approach, we conduct experiments focusing on two key aspects:  
1) The robustness of the method in handling open-set instructions and open-set objects.  
2) The alignment between the generated trajectory and the given instruction.

\subsection{Navigation under Open-Set Instruction and Objects}
\textbf{Task Description.} To evaluate the robustness of our proposed zero-shot framework in real-world outdoor navigation tasks given free-form language instructions, for each scene from SemanticKITTI sequence 05, we consider two categories of navigation tasks.

1) \textbf{Move-to-object task}.  
In this category, the robot must navigate from its current position to a target object. We progressively relax the constraints on the object description in the instruction to evaluate the algorithm's performance on open-set objects:
\begin{itemize}
    \item \textbf{Specific} object navigation: The instruction explicitly specifies the target object, and the environment contains only one such object, for example, "Go to the red car."
    \item \textbf{Ambiguous} object navigation: The scene contains multiple similar objects, requiring the robot to resolve ambiguity using multimodal information. Instructions may include spatial references, appearance, or other subtle cues, for example, "Go to the second car on your left" or "Wait for me by the tree near the stop sign."
\end{itemize}

\begin{table}[t!]
    \centering
    \renewcommand{\arraystretch}{1.2} 
    \begin{tabular}{llcc|cc}
        \toprule
        \multirow{2}{*}{\textbf{Task-Type}} & \multirow{2}{*}{\textbf{Subcategories}} & \multicolumn{2}{c|}{\textbf{LLM-TG}\cite{kwon2024language}} & \multicolumn{2}{c}{\textbf{OpenNav}} \\ 
        \cmidrule(lr){3-4} \cmidrule(lr){5-6}
        & & \textbf{NE} & \textbf{SR} & \textbf{NE} & \textbf{SR} \\ 
        \midrule
        \multirow{2}{*}{Move to obj.} & Specificity    & 1.44 & 43/50 & \textbf{1.01} & \textbf{45/50} \\
                                      & Ambiguity    & 7.40 & 13/50 & \textbf{1.63} & \textbf{42/50} \\ 
        \midrule
        High-level & Reasoning    & 5.90 & 23/50 & \textbf{2.40} & \textbf{40/50} \\ 
        \midrule
        \multicolumn{2}{l}{\textbf{Total}} & 4.91 & 53\% & \textbf{1.68} & \textbf{84\%} \\ 
        \bottomrule
    \end{tabular}
    \caption{Performance comparison of LLM-TG and OpenNav. SR: Success Rate, NE: Navigation Error.}
    \label{tab:performance1}
    \vspace{-1.2em}
\end{table}

2) \textbf{High-level language instruction tasks.}  
This category assesses the flexibility of human-robot interaction, where the instruction does not explicitly mention the target object or cannot be resolved using a single object caption. The model must understand the task, perform \textbf{reasoning}, and integrate environmental observations to generate the trajectory. Examples include: ``Throw away the trash'', where the model must infer that a trash bin is the most relevant target; ``Go straight first, then turn right into the courtyard.``, where the model must decompose the instruction into sequential navigation steps, and then selects the most relevant environmental information to generate a trajectory.

\textbf{Evaluation Metrics.} To evaluate the effectiveness of our system, we conduct experiments across various scenarios and report the following quantitative metrics.  
\begin{itemize}
    \item Success Rate (SR): A trajectory succeeds if its final position is within 1 meter of the ground truth.
    \item Navigation Error (NE): The average Euclidean distance between the estimated and ground-truth endpoints.
\end{itemize}

\textbf{Compared Baseline.} We compare our approach with LLM-TG, a framework proposed by Kwon et al.\cite{kwon2024language}. In LLM-TG, the observational information relies entirely on data provided by the object detector, including object captions, coordinates, sizes, and the nearest reachable points on each drivable region to the target. In contrast, our method leverages multimodal inputs. To ensure a fair comparison, both frameworks use ChatGPT-4o as the decision-making model, differing only in the modality of the input provided.

\textbf{Results and Analysis.} The experimental results presented in Table~\ref{tab:performance1} indicate that when no ambiguity is present in the instruction, the performance of LLM-TG and our multimodal framework remains similar, both achieving a 100\% success rate. However, in scenarios where instruction is ambiguous, the performance of the text-only approach significantly degrades, while our method maintains a high success rate.
\begin{figure}[t] 
	\centering
	\includegraphics[width=0.8\linewidth]{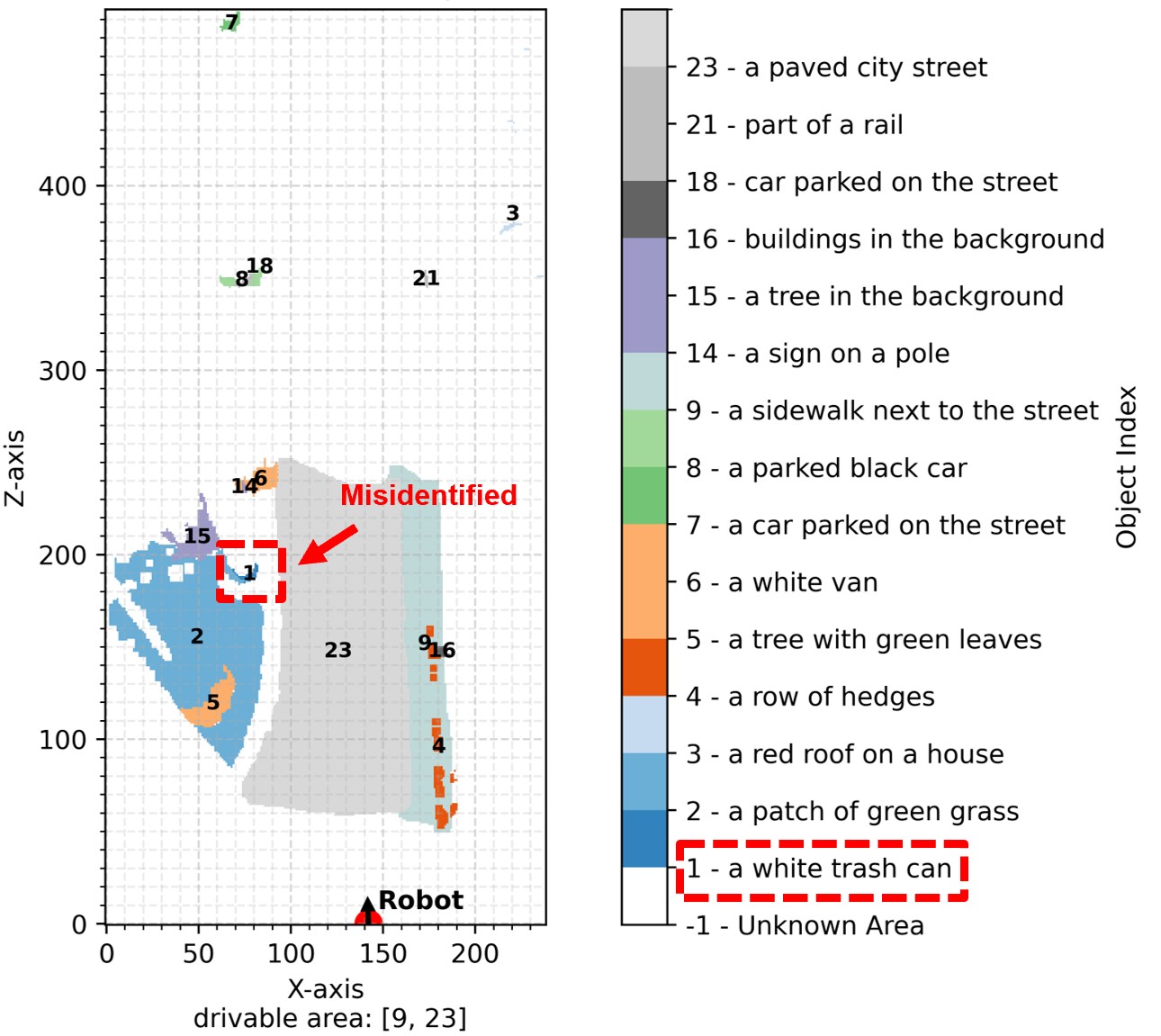}
	\caption{A 2D semantic map with captions and corresponding numerical labels generated by OVPS, where a cylindrical building with label-1 was misidentified as a trash can. The corresponding scenario can be seen in Fig. \ref{fig:sample}, case three.}
	\label{fig:detection_error}
    \vspace{-1.25em}
\end{figure}
Failure cases of LLM-TG reveal that LLM-TG failures are primarily attributed to two key factors: 1) Dependence on object detector accuracy - errors in object detection directly impact navigation success, and 2) Granularity of captioned information - limited descriptive details in captions reduce disambiguation capability.
For example, the instruction "Go to the cylindrical building" leads to failure in LLM-TG because: The object detector mistakenly classifies the cylindrical structure as a trash bin (see Fig. \ref{fig:detection_error}). The method, relying solely on caption information, is unable to identify the correct target. In contrast, our method first identifies the numerical label of the cylindrical building in the annotated image, then verifies the corresponding object caption (trash bin) and is aware of the misclassification from OVPS. The final trajectory is computed based on the nearest accessible point on the ground associated with the corrected object label, successfully completing the task, as shown in Fig.~\ref{fig:sample}. Another failure case of LLM-TG is shown in Fig.~\ref{fig:sample}, where the instruction ``Go to the house on your right'' is given. Multiple houses exist in the environment. The robot must correctly interpret spatial references to determine the final target. Our approach outperforms the text-only method by leveraging scene understanding from the annotated image. In addition, our method exhibits a significant advantage in handling high-level language instructions compared to LLM-TG, as shown in the test results provided in Table \ref{tab:performance1}. This validates our initial hypothesis that a unified perception and planning framework is crucial for robot navigation. Robots should plan trajectories based on what they see, rather than solely relying on textual descriptions. 

\subsection{Language Guided Zero-Shot Trajectory Synthesis}
\begin{figure*}[t]
    \centering
    \begin{subfigure}{0.65\linewidth}
        \centering
        \includegraphics[width=\linewidth]{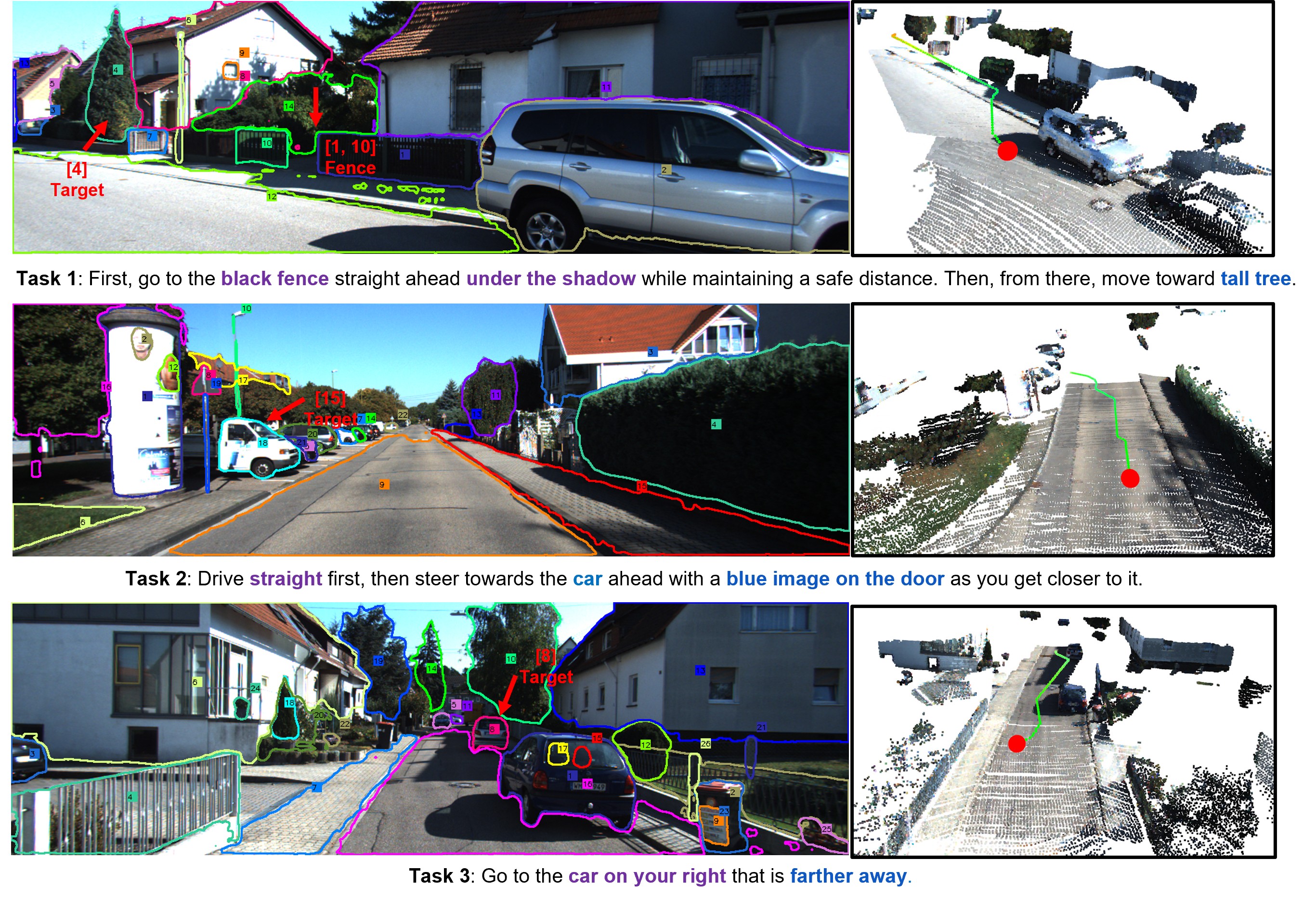}
        \caption{Tasks and the generated trajectories with OpenNav}
        \label{fig:tasks}
    \end{subfigure}
    \hfill
    \begin{subfigure}{0.32\linewidth}
        \centering
        \includegraphics[width=\linewidth]{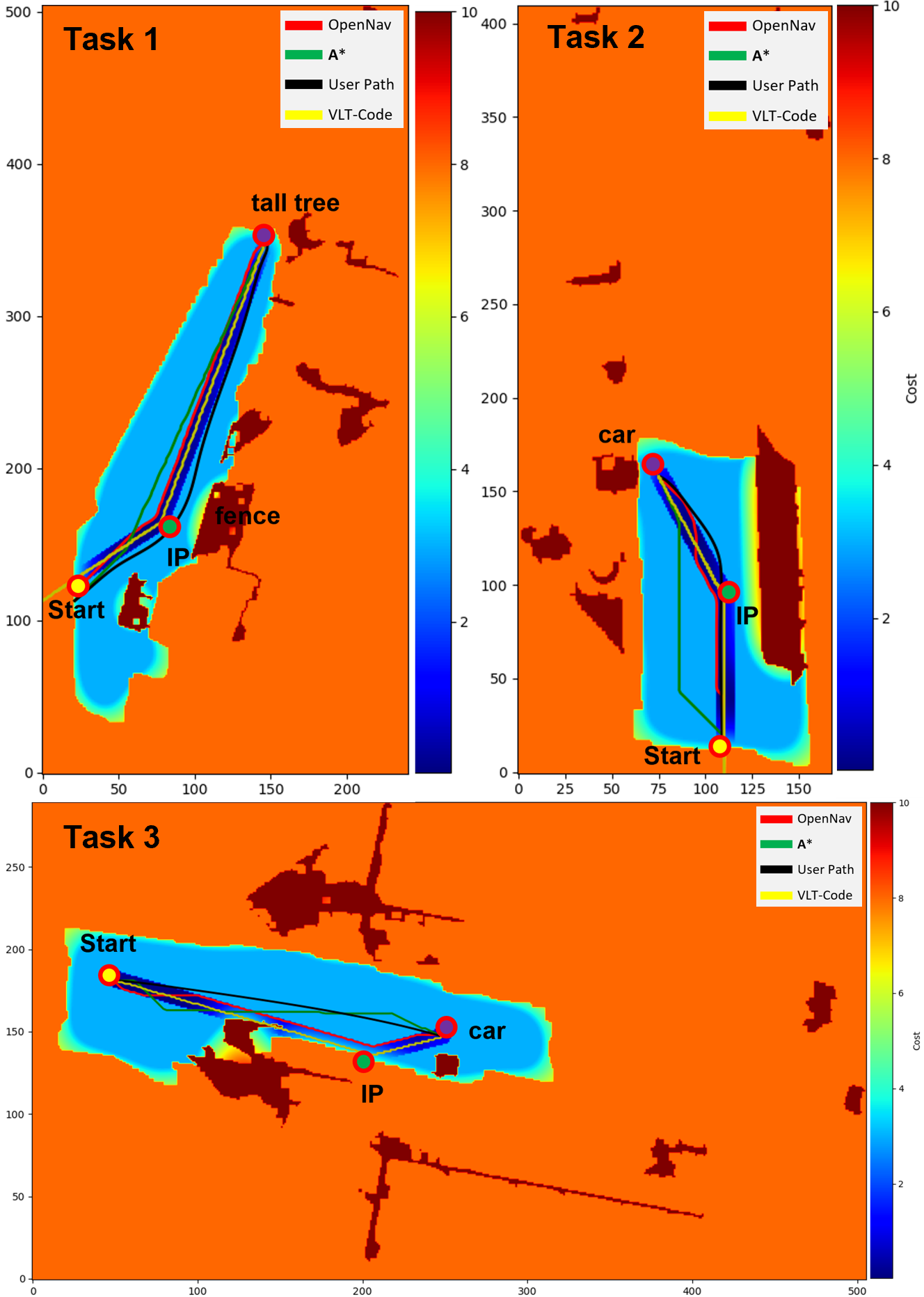}
        \caption{Value maps}
        \label{fig:valuemap}
    \end{subfigure}
    \caption{Selected examples demonstrating how OpenNav utilizes Value Maps to generate task-aligned and geometry-compliant navigation trajectories: (a) User-specified tasks and trajectories generated by OpenNav; b) Value maps illustrating trajectories generated by different algorithms based on the given tasks.}
    \label{fig:comparison}
    \vspace{-1.6em}
\end{figure*}
This section evaluates how well the trajectories generated with OpenNav align with human-annotated trajectories based on the language instructions. To assess this alignment, we use trajectory shape similarity metrics, where higher similarity indicates better performance in language-guided trajectory synthesis.  This experiment serves as an ablation study, comparing three trajectory generation methods:
\begin{itemize}
    \item A*: Generates collision-free trajectories solely based on the occupancy map, using only the start and goal points.
    \item VLT-Code: A baseline where an MLLM directly generates trajectories via code, selecting start, intermediate (IP), and end points without a value map, relying solely on its spatial reasoning (see Section~\ref{sec: VLT}).
    \item OpenNav: our proposed full-version VLN framework, integrating MLLM with geometry-compliant trajectory refinement, as described in Section \ref{sec: OpenNav}.
\end{itemize}

For evaluation, users provide a language instruction and then manually draw their preferred trajectory on the map, serving as the ground truth for assessing the quality of generated trajectories. To comprehensively evaluate the similarity between generated and user-defined trajectories, we use both Fréchet Distance\cite{alt1995computing} and Normalized Dynamic Time Warping (NDTW)\cite{ilharco2019general}. Fréchet Distance strictly preserves trajectory shape, making it sensitive to deviations, while NDTW accounts for speed variations by allowing flexible temporal alignment. A lower Fréchet Distance indicates higher spatial similarity, whereas a higher NDTW score reflects better overall alignment. 

The experimental results are shown in Table~\ref{tab:performance2}. The A* algorithm excels in collision avoidance, with no collisions recorded across the ten selected task scenarios. However, its trajectory similarity performance is suboptimal, which is expected since A* focuses on finding the shortest trajectory between the start and goal while lacking flexibility to shape trajectories based on task requirements. In contrast, the VLT-Code method improves trajectory similarity compared to A*, yet its reliance on pre-trained MLLMs, which lack depth and geometry information during training, results in poor performance in spatial collision avoidance. Our proposed \textbf{OpenNav} demonstrates a clear improvement in both trajectory similarity and collision avoidance.
\begin{table}[t!]
	\caption {\label{tab:performance2} Performance in Trajectory Synthesis} 
    \vspace{-0.3em}
	\centering
	\begin{tabular}{cccc}
		\toprule
		\textbf{Methods} & \textbf { Fréchet Distance} $\downarrow$  & \textbf{NDTW} $\uparrow$ & \textbf{Collision} $\downarrow$\\
		\midrule
		A* & 24.20 & 0.08 & 0/30  \\
		VLT-Code & 17.53 & 0.16 & 16/30  \\
		OpenNav & \textbf{12.60} & \textbf{0.38} & \textbf{2/30}  \\
		\bottomrule
	\end{tabular}
    \vspace{-1.7em}
\end{table}
Fig.~\ref{fig:tasks} presents three representative tasks. In \textbf{Task 1}, the instruction is: \textit{``Drive straight to the black fence under the shadow, then proceed toward the tall tree."} This requires the model to decompose the task into two steps: first, identifying an intermediate waypoint near the fence based on the current position and orientation, and second, navigating from that point to the final goal. The user-defined trajectory, shown as a black line in Fig.~\ref{fig:valuemap}, serves as ground truth. The value map in \textbf{Task 1} reveals that A* (green) disregards the intermediate waypoint and directly heads toward the tall tree, whereas both VLT-Code (yellow) and OpenNav (red) successfully incorporate the intermediate waypoint, resulting in trajectories more aligned with the user-defined trajectory.
Similarly, in \textbf{Task 2}, where the instruction is: \textit{``Drive straight first, then head toward the white car with a blue pattern on the door."}, both VLT-Code and OpenNav effectively follow the intended trajectory. However, VLT-Code sometimes generates trajectories that align with the task intent but, due to its limited spatial reasoning, may lead to collisions. As illustrated in Fig.~\ref{fig:valuemap} for \textbf{Task 3}, the VLT-Code trajectory passes through an occupied region where a vehicle is present. For OpenNav, the results clearly show that it successfully integrates MLLM with the advantages of geometry-based spatial constraints, allowing it to balance trajectory adherence and collision avoidance, ensuring a low collision rate while closely following user-preferred paths. 
Due to space constraints, we provide more comprehensive experimental results and real-robot demonstrations on our project website.

\section{CONCLUSIONS}
This work introduces OpenNav, a zero-shot VLN framework that successfully translates the powerful language reasoning, high-level planning, and code-generation capabilities of MLLMs into executable robot actions by generating navigation trajectories. Without any training data or scene-specific prompt engineering, our approach demonstrates strong generalization to open-set instructions and open-set objects in challenging unstructured scenarios.
Additionally, we propose a task-agnostic prompt strategy, and a multimodal interface design between perception system and MLLM to successfully address the granularity issues of environmental perception, particularly for free-form language instructions, while also exhibiting robustness against misdetections from detectors. Our work combines the strengths of MLLMs with the classical planning algorithm, ensuring that the generated trajectories align with language instructions while adhering to environmental geometric constraints to enable collision avoidance.
We validate our approach on both AVDs and real-world wheeled robot deployments under diverse scenarios and task types, achieving consistently strong performance. In the future, we aim to extend this method to more complex and dynamic environments.
\bibliographystyle{IEEEtran}

\end{document}